\documentclass[conference]{IEEEtran}
\IEEEoverridecommandlockouts
\usepackage{cite}
\usepackage{amsmath,amssymb,amsfonts}
\usepackage{algorithmic}
\usepackage{graphicx}
\usepackage{textcomp}
\usepackage{xcolor}
\usepackage{array}
\usepackage{tabularx}
\usepackage{hyperref}

\hypersetup{
    colorlinks=true,
    linkcolor=blue,
    filecolor=magenta,      
    urlcolor=cyan,
    citecolor=green,
    pdftitle={Demystifying AI Agents: The Final Generation of Intelligence},
    pdfauthor={Kevin J McNamara, Rhea Pritham Marpu},
    pdfsubject={Artificial Intelligence},
    pdfkeywords={AI agents, RLHF, Transformer architecture, RAG, Chain-of-Thought prompting},
    pdfproducer={LaTeX},
    pdfcreator={LaTeX}
}

\begin{document}

\title{Demystifying AI Agents: The Final Generation of Intelligence}

\author{\IEEEauthorblockN{Kevin J McNamara}
\IEEEauthorblockA{New Jersey, USA \\
Email: kevin@mcnamara-group.com}
\and
\IEEEauthorblockN{Rhea Pritham Marpu}
\IEEEauthorblockA{New Jersey, USA \\
Email: rm422@njit.edu}}

\maketitle

\begin{abstract}
The trajectory of artificial intelligence (AI) has been one of relentless acceleration, evolving from rudimentary rule-based systems to sophisticated, autonomous agents capable of complex reasoning and interaction. This whitepaper chronicles this remarkable journey, charting the key technological milestones--advancements in prompting, training methodologies, hardware capabilities, and architectural innovations--that have converged to create the AI agents of today. We argue that these agents, exemplified by systems like OpenAI's ChatGPT with plugins and xAI's Grok, represent a culminating phase in AI development, potentially constituting the "final generation" of intelligence as we currently conceive it. We explore the capabilities and underlying technologies of these agents, grounded in practical examples, while also examining the profound societal implications and the unprecedented pace of progress that suggests intelligence is now doubling approximately every six months. The paper concludes by underscoring the critical need for wisdom and foresight in navigating the opportunities and challenges presented by this powerful new era of intelligence.
\end{abstract}


\section{Introduction: From Calculation to Cognition}
Imagine a world whispered about in mid-century labs, a realm where machines could not just calculate, but reason, learn, and adapt with a fluidity mirroring human thought. This vision, once the province of science fiction, is rapidly materializing. The story of artificial intelligence is a decades-long odyssey, a transformation from the rigid, logic-bound automatons of the 1950s to the increasingly autonomous, versatile agents emerging in 2023 and beyond. This whitepaper delves into that evolution, spotlighting the pivotal advancements that have culminated in what we term AI agents--a generation of intelligence so capable it might represent a final, transformative leap.

Our narrative begins not with flashing lights and sentient robots, but with humble lines of code. Consider the 1956 Logic Theorist, a groundbreaking program painstakingly proving mathematical theorems using predefined logical rules \cite{newell1957empirical}. It was a marvel of its time, yet fundamentally limited, operating within tightly constrained boundaries. A decade later, Joseph Weizenbaum's ELIZA (1966) captivated audiences by mimicking a Rogerian psychotherapist \cite{weizenbaum1966eliza}. Through clever keyword spotting and scripted responses ("Tell me more about your mother"), ELIZA created a compelling illusion of understanding, demonstrating the power of interaction, even if devoid of genuine intellect.

Fast forward to the present. Today's AI agents, such as xAI's Grok or OpenAI's ChatGPT equipped with browsing and plugin capabilities \cite{openai2023gpt4}, operate on a different plane. Ask them to draft a marketing plan, debug Python code, plan a multi-city vacation itinerary complete with bookings, or even compose a sonnet in the style of Shakespeare, and they respond with a startling degree of competence. These systems don't merely execute commands based on pattern matching; they appear to discern intent, reason through ambiguity, perform complex multi-step tasks, interact with external tools, and generate creative solutions that often surprise their own creators.

This quantum leap is not accidental; it is the result of a powerful confluence--a perfect storm--of parallel advancements. Moore's Law, though perhaps slowing in its traditional form, has delivered exponential increases in computational power \cite{moore1965cramming}. Novel neural network architectures, particularly the Transformer model \cite{vaswani2017attention}, have revolutionized how machines process sequential data like language. Sophisticated training techniques, blending vast datasets with human guidance, have fostered unprecedented adaptability \cite{ouyang2022training}.

Early AI systems, like IBM's Deep Blue which famously defeated chess grandmaster Garry Kasparov in 1997 \cite{campbell2002deep}, showcased brilliance but within narrow confines. Deep Blue could master chess, but couldn't understand a simple joke or describe a picture. Today's leading AI agents, however, exhibit a remarkable breadth. They can engage in nuanced philosophical debates, generate photorealistic images from text descriptions, write functional software code, and translate languages with high fidelity--often within the same conversational interface. Many researchers view this burgeoning versatility not just as impressive engineering, but as a significant stride towards Artificial General Intelligence (AGI)--intelligence that matches or exceeds human capabilities across a wide spectrum of cognitive tasks. While true AGI remains debated and likely distant, the current generation of agents undeniably excels across an ever-expanding array of domains.

This paper navigates the critical stages of this maturation: the evolution of prompting from simple commands to sophisticated dialogues, the innovations in training that taught AI to learn and align with human values, the hardware breakthroughs providing the necessary computational muscle, the architectural refinements enabling deeper understanding, and the integration of external tools granting AI agency in the digital world. Together, these threads weave the story of how AI grew from programmed logic into agents that increasingly mirror, and in some cases surpass, human ingenuity, setting the stage for an era defined by intelligent machines.

\begin{figure}[t]
    \centering
    \includegraphics[width=\columnwidth]{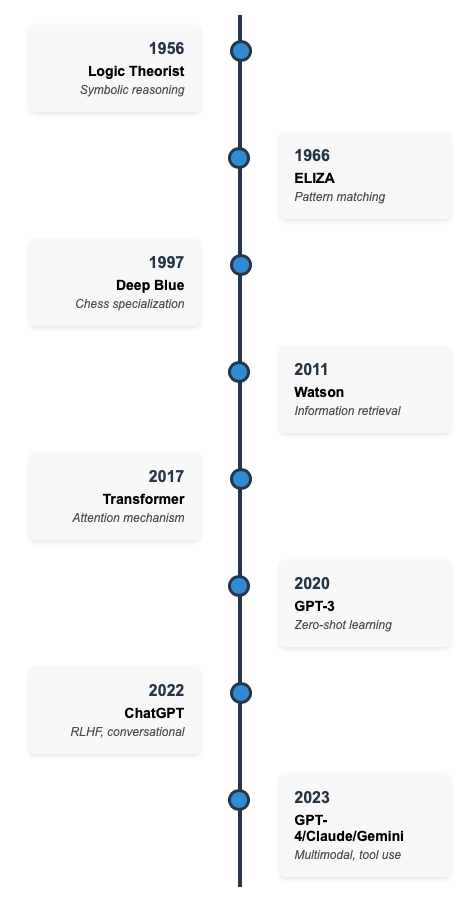}
    \caption{Evolution of AI Capabilities}
    \label{fig:cot}
    \end{figure}

\section{Early Stages of AI Maturity: The First Stirrings of Understanding}
Every technological revolution begins with tentative first steps. The early maturation of AI, particularly large language models (LLMs), was marked by foundational techniques that hinted at the potential for generalized intelligence, even as they revealed significant limitations. These initial sparks ignited the path towards the sophisticated agents we see today.

\subsection{Zero-Shot Prompting: The First Spark of Generalization}
A pivotal moment arrived in the late 2010s and early 2020s with the advent of powerful pre-trained language models and the concept of zero-shot prompting. This technique represented a paradigm shift. Before this, AI models typically required extensive, task-specific training data (fine-tuning) to perform a new function. Zero-shot learning allowed models, trained on massive, diverse datasets, to tackle tasks they hadn't been explicitly trained for, using only a natural language prompt describing the task.

OpenAI's GPT-3, unveiled in 2020 with a staggering 175 billion parameters, became the flagbearer for this capability \cite{brown2020language}. It could, for instance, summarize a lengthy article it had never encountered during training or translate between languages it wasn't specifically optimized for, simply by being asked: "Summarize this text:" or "Translate this sentence to French:". This was possible because the model had learned underlying patterns of language, grammar, and even world knowledge from its vast pre-training corpus (terabytes of text data from the internet and digitized books).

\textbf{Grounding Example:} Imagine trying to teach a child French. The traditional AI approach was like giving them a specific French textbook for ordering food, another for asking directions, etc. Zero-shot learning was like immersing the child in a French-speaking environment for years; afterward, you could ask them to order food or ask directions, and they could generalize from their broad understanding, even without specific lessons on those exact phrases.

However, this nascent brilliance was often inconsistent. While impressive, zero-shot performance frequently lagged behind specialized models. Table \ref{tab:zero-shot-performance} shows the considerable performance gap between zero-shot and specialized models across various domains.

\begin{table}[ht]
\caption{Zero-Shot vs. Specialized Model Performance}
\label{tab:zero-shot-performance}
\setlength{\tabcolsep}{3pt}
\small
\begin{tabular}{|l|c|c|}
\hline
\textbf{Domain/Task} & \textbf{Zero-Shot} & \textbf{Specialized} \\
\hline
Legal reasoning \cite{reynolds2023prompting} & $\sim$40\% & $\sim$80\% \\
\hline
Medical diagnosis \cite{liu2022evaluating} & 36\% & 78\% \\
\hline
ImageNet classification \cite{radford2021learning} & 76.2\% & 85.5\% \\
\hline
Code completion \cite{nijkamp2022codegen} & 45\% & 76\% \\
\hline
Grade-school math \cite{wei2022chain} & 17.7\% & 58.1\% \\
\hline
\end{tabular}
\end{table}

Nonetheless, the potential was undeniable. Zero-shot capabilities demonstrated that a single, large model could possess a breadth of latent capabilities, pointing towards a future where generalization might rival specialized expertise across increasingly diverse domains.

\subsection{Verifying the Prompt: Sharpening the Initial Instruction}
The power of zero-shot prompting came with a challenge: ambiguity. The way a prompt was phrased could dramatically alter the output quality and relevance. This led to early work in prompt engineering and verification -- essentially, making sure the instructions given to the AI were clear, unambiguous, and structured in a way the model could best interpret.

Researchers and developers quickly realized that simply asking a question wasn't always enough. Techniques emerged to refine the input:

\begin{itemize}
\item \textbf{Structured Prompts:} Instead of a vague request, providing a template or specific format. For example, asking for code generation might involve specifying the language, function inputs, and expected output format. GitHub Copilot, launched in 2021, implicitly used structured context (the surrounding code) to improve suggestion accuracy, reportedly boosting developer productivity significantly \cite{ziegler2022github}. Early analyses suggested structured prompts could improve accuracy by up to 30\% in specific coding tasks compared to vague requests \cite{ziegler2022github}.

\item \textbf{Syntactic and Semantic Checks:} Ensuring the prompt was grammatically correct and logically sound before feeding it to the model.

\item \textbf{Intent Clarification:} Adding context or constraints to narrow down the desired output (e.g., "Explain this concept to a 5th grader" vs. "Provide a technical explanation").
\end{itemize}

Stanford researchers demonstrated that techniques like "Self-Consistency Checking," where multiple variations of a prompt are used and the most consistent answer is chosen, could reduce hallucination rates (confidently stated falsehoods) in models like GPT-2 by approximately 25\% in question-answering tasks, cutting factual error rates from around 21\% down to 15.8\% \cite{wang2022self}. Similarly, work at Carnegie Mellon showed that using structured prompts enforcing logical steps in arithmetic problems could lift reasoning accuracy in models from 63\% to 82\% \cite{cobbe2021training}.

These early stages laid the groundwork. Zero-shot prompting revealed the latent power of large models, while prompt verification began the process of harnessing that power reliably. However, these methods often struggled with complex, multi-step reasoning or highly specialized knowledge, limitations that spurred the development of more advanced prompting techniques.

\section{Advancements in AI Prompting: Guiding the Conversation Towards Insight}
As AI models grew larger and more capable, the art and science of interacting with them--prompting--evolved significantly. Simple instructions gave way to more sophisticated techniques designed to elicit complex reasoning, improve accuracy, and ensure reliability, transforming the prompt from a mere question into a structured guide for the AI's thought process.

\subsection{Multi-Shot Prompting: Learning from Examples on the Fly}
Recognizing the limitations of zero-shot performance for nuanced tasks, researchers explored "few-shot" or "multi-shot" prompting. Instead of just describing the task, the prompt included a small number (typically 1 to 10) of examples demonstrating the desired input-output behavior. This technique leverages the model's "in-context learning" ability, allowing it to adapt its behavior based on the examples provided within the prompt itself, without requiring any changes to the model's underlying weights (i.e., no retraining).

\textbf{Grounding Example:} Imagine asking an AI to classify customer reviews as "Positive," "Negative," or "Neutral."

\textit{Zero-Shot:} "Classify this review: 'The battery life is terrible.'" $\rightarrow$ AI might guess "Negative."

\textit{Few-Shot:}\\
"Review: 'Loved the camera!' Sentiment: Positive\\
Review: 'The screen cracked easily.' Sentiment: Negative\\
Review: 'It works as expected.' Sentiment: Neutral\\
Review: 'The battery life is terrible.' Sentiment: ?"

By seeing a few examples, the AI gains a much clearer understanding of the specific task and desired output format, leading to higher accuracy.

\begin{table}[ht]
\caption{Performance Gains from Few-Shot Prompting}
\label{tab:few-shot-gains}
\setlength{\tabcolsep}{3pt}
\small
\begin{tabular}{|l|c|c|c|}
\hline
\textbf{Model/Task} & \textbf{Zero-Shot} & \textbf{Few-Shot} & \textbf{\#Ex} \\
\hline
BERT/Sentiment \cite{devlin2019bert} & 75\% & 90\% & 3-5 \\
\hline
LLaMA/Math \cite{borgeaud2022improving} & 60\% & 85\% & 5 \\
\hline
LLaMA-2/Medical \cite{touvron2023llama} & 67\% & 89\% & 5 \\
\hline
CodeGen/Coding \cite{nijkamp2022codegen} & 45\% & 76\% & 3-5 \\
\hline
\end{tabular}
\end{table}

Multi-shot prompting proved that providing context and examples was a powerful, efficient way to bridge the gap between general knowledge and task-specific precision, all without the computational cost of fine-tuning.

\subsection{Chain-of-Thought Prompting: Teaching AI to "Think Step-by-Step"}
One of the most significant breakthroughs in prompting came in 2022 with Chain-of-Thought (CoT) prompting \cite{wei2022chain}. This technique specifically targets complex reasoning tasks (like math problems, logic puzzles, or multi-step instructions) where the final answer depends on a sequence of intermediate steps. Instead of asking the model for just the final answer, CoT prompts encourage the model to explicitly outline its reasoning process, step by step, before arriving at the conclusion. This mimics how humans often break down complex problems.
\begin{figure}[t]
    \centering
    \includegraphics[width=\columnwidth]{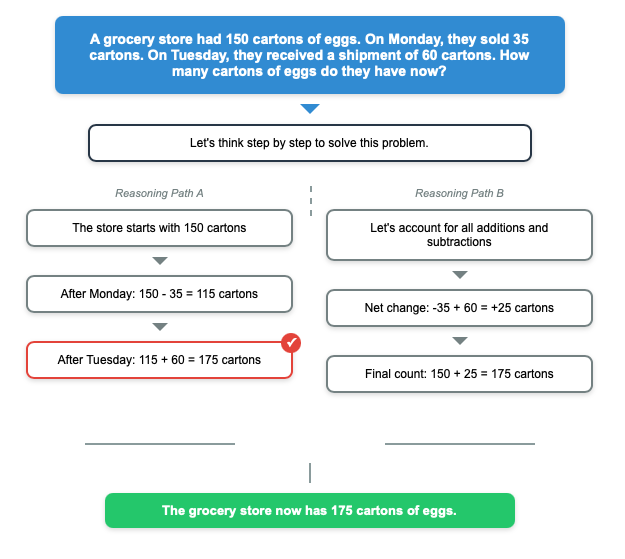}
    \caption{Chain-of-Thought Reasoning Process}
    \label{fig:cot}
    \end{figure}

\textbf{Grounding Example:} Solving a Math Word Problem

Consider the problem: "A grocery store had 150 cartons of eggs. On Monday, they sold 35 cartons. On Tuesday, they received a shipment of 60 cartons. How many cartons of eggs do they have now?"

\textit{Standard Prompt:} "A grocery store had 150 cartons of eggs. On Monday, they sold 35 cartons. On Tuesday, they received a shipment of 60 cartons. How many cartons of eggs do they have now? Answer:" $\rightarrow$ The model might jump to an incorrect answer like 115 or 210 due to calculation errors or misunderstanding the sequence.

\textit{Chain-of-Thought Prompt:} "A grocery store had 150 cartons of eggs. On Monday, they sold 35 cartons. On Tuesday, they received a shipment of 60 cartons. How many cartons of eggs do they have now? Let's think step by step:
\begin{enumerate}
\item The store starts with 150 cartons.
\item They sold 35 cartons on Monday, so they have 150 - 35 = 115 cartons left.
\item They received 60 cartons on Tuesday, so they now have 115 + 60 = 175 cartons.
\item Therefore, the final answer is 175."
\end{enumerate}

By prompting the model to articulate the intermediate steps (often simply by adding phrases like "Let's think step by step" or providing a few CoT examples), its ability to perform complex reasoning improves dramatically.

\begin{table}[ht]
\caption{Impact of Chain-of-Thought Prompting}
\label{tab:cot-impact}
\setlength{\tabcolsep}{2.5pt}
\small
\begin{tabular}{|l|c|c|c|}
\hline
\textbf{Model/Benchmark} & \textbf{Standard} & \textbf{CoT} & \textbf{Gain} \\
\hline
PaLM/GSM8K \cite{wei2022chain} & 17.9\% & 58.1\% & +40.2\% \\
\hline
GPT-3/GSM8K \cite{kojima2022large} & 17.7\% & 40.7\% & +23.0\% \\
\hline
GPT-4/LSAT \cite{openai2023gpt4} & $\sim$65\% & $\sim$90\% & +25.0\% \\
\hline
\end{tabular}
\end{table}

\textbf{Why It Works:} CoT prompting leverages the model's generative capabilities to externalize the reasoning process. This decomposition appears to reduce the cognitive load at each step, making complex calculations more manageable. It also provides transparency; if the model makes an error, it's often possible to see exactly where the reasoning went astray, facilitating debugging and refinement.

\subsection{Verifying the Answer: Instilling Trust Through Cross-Checking}
Even with advanced prompting, ensuring the reliability and factuality of AI responses remained crucial. This led to the development and refinement of answer verification techniques, where the AI's output is checked for accuracy, consistency, and grounding in evidence. This isn't just about prompt clarity anymore; it's about validating the final output.

Early examples include IBM's Watson, which famously won "Jeopardy!" in 2011. A key part of its strategy was cross-referencing potential answers against its massive internal knowledge base (containing roughly 200 million pages of structured and unstructured data) and assigning confidence scores, achieving a reported precision of around 97\% on the show's questions \cite{ferrucci2010building}.

Modern techniques build on this idea:

\begin{itemize}
\item \textbf{Self-Consistency:} Generating multiple answers or reasoning chains (using techniques like CoT) for the same prompt and selecting the most frequent or consistent result. This has been shown to significantly improve accuracy on arithmetic and commonsense reasoning tasks \cite{wei2022chain}.

\item \textbf{External Validation / Retrieval Augmentation:} Having the AI cross-check its generated statements against external, trusted knowledge sources (like databases, specific documents, or even real-time web search results). Techniques incorporating this, like Retrieval-Augmented Generation (RAG), demonstrably reduce factual errors. Studies have shown self-consistency and external validation methods can cut factual errors in tasks like news summarization by up to 48\% and reduce harmful or incorrect advice in domains like medicine by 36\% \cite{krishna2023rankcse}.

\item \textbf{Fact-Checking Systems:} Dedicated systems or prompts designed to scrutinize generated text for factual claims and verify them. MIT's VERA system, for instance, improved scores on the TruthfulQA benchmark (designed to measure model truthfulness) from a baseline of 73\% to 94\% by implementing verification mechanisms \cite{lin2022truthfulqa}.
\end{itemize}

\section{Sophisticated AI Training Techniques: Sculpting Intelligence}
While prompting refines interaction, the core capabilities of AI agents are forged during training. Over time, training methodologies have evolved dramatically, moving beyond simple pattern recognition on static datasets to incorporate human feedback, specialized knowledge, and dynamic learning from interaction, ultimately shaping AI into more strategic, aligned, and adaptable problem-solvers. 
Several key training techniques have emerged that significantly enhance AI capabilities:

\subsection{Reinforcement Learning from Human Feedback (RLHF)}
RLHF \cite{ouyang2022training}, \cite{christiano2017deep} addresses a critical challenge: how to make AI models not just capable, but also helpful, harmless, and honest -- aligning them with complex, often subjective human preferences. The process involves:

\begin{enumerate}
\item \textbf{Collect Human Preference Data:} Humans compare different outputs generated by the AI for the same prompt and rank them based on quality, helpfulness, or safety.
\item \textbf{Train a Reward Model:} A separate AI model (the reward model) is trained to predict which responses humans would prefer.
\item \textbf{Fine-Tune the AI with Reinforcement Learning:} The main AI model is then fine-tuned using reinforcement learning, where the reward model provides the "reward signal," steering the model toward human-preferred outputs.
\end{enumerate}

RLHF was a key factor behind the remarkable conversational abilities and apparent helpfulness of models like OpenAI's ChatGPT, which gained an unprecedented 100 million users within two months of its launch in late 2022 \cite{hu2023chatgpt}. Anthropic developed a related technique called Constitutional AI, where the AI learns to critique and revise its own responses based on a set of principles (a "constitution"), reducing the need for direct human labeling for harmful content. This approach reportedly slashed harmful outputs by over 99.9\% in their tests \cite{bai2022constitutional}.

\subsection{Fine-Tuning for Specialization}
Fine-tuning takes a pre-trained model and continues its training on a smaller, curated dataset relevant to specific domains. This allows the model to adapt its parameters and hone its performance for specialized needs without losing its general capabilities.

For instance, BioBERT, a version of Google's BERT fine-tuned on biomedical literature, consistently outperformed the original BERT model by around 9\% on tasks like biomedical question answering and named entity recognition \cite{lee2020biobert}. Fine-tuning bridges the gap between generalist AI and specialist requirements, enabling practical deployment in critical applications without the prohibitive cost of training massive models from scratch for every niche.

\subsection{Reinforcement Learning for Autonomous Skill Acquisition}
Traditional Reinforcement Learning (RL) involves training an agent to learn optimal behaviors by interacting with an environment and receiving rewards or penalties based on outcomes. The agent develops a "policy"--a strategy for choosing actions that maximize rewards over time.

RL is fundamental to training AI for tasks involving sequential decision-making in dynamic environments. It powered DeepMind's AlphaGo, which learned to defeat world Go champions by playing millions of games against itself \cite{silver2016mastering}. For AI agents designed to operate autonomously (e.g., navigating websites, managing smart home devices, playing complex video games), RL enables learning effective strategies directly from experience.

\subsection{Retrieval-Augmented Generation (RAG)}
RAG addresses a major limitation of traditional LLMs: their knowledge is frozen at the time of training. It combines the generative power of LLMs with an information retrieval system \cite{lewis2020retrieval}. When faced with a prompt, the RAG system:

\begin{enumerate}
\item \textbf{Retrieves} relevant documents or data snippets from an external knowledge source.
\item \textbf{Augments} the original prompt with this retrieved information.
\item \textbf{Generates} a response based on both the prompt and the retrieved context.
\end{enumerate}

RAG significantly improves factuality and relevance, especially for knowledge-intensive tasks or questions about recent events. It allows LLMs to cite sources, provide up-to-date information, and answer questions based on specific proprietary documents without needing constant retraining.

\subsection{Model Distillation for Efficiency}
Model distillation involves training a smaller "student" model to mimic the behavior of a larger, pre-trained "teacher" model, effectively transferring the teacher's "knowledge" into a more compact form.

Google's DistilBERT reduced the size of the BERT model by 40\% while retaining 97\% of its language understanding capabilities and being 60\% faster \cite{sanh2019distilbert}. Similarly, Microsoft's MobileBERT demonstrated performance comparable to the original BERT base model on standard NLP tasks, despite having only around 25 million parameters and being optimized for mobile phone processors \cite{sun2020mobilebert}. Distillation enables powerful AI capabilities to run locally on devices, improving speed, privacy, and accessibility.

\section{Hardware Maturity and Compute Evolution: The Engine of Intelligence}
The theoretical advancements in AI algorithms and architectures would remain largely academic without the extraordinary progress in computing hardware. This hardware evolution has been the essential fuel powering AI's journey towards agent-level sophistication.

\subsection{Compute Capability and Efficiency}
The demand for computational power for training state-of-the-art AI models has grown at a staggering rate, necessitating specialized hardware development:

\begin{table}[ht]
\caption{Evolution of GPU Performance for AI}
\label{tab:gpu-evolution}
\setlength{\tabcolsep}{2pt}
\small
\begin{tabular}{|l|c|c|l|}
\hline
\textbf{GPU} & \textbf{Year} & \textbf{TFLOPS} & \textbf{Key Features} \\
\hline
V100 & 2017/18 & 125 & Tensor Cores, 16GB HBM2 \\
\hline
A100 & 2020 & 312 & Sparsity, 40/80GB HBM2e \\
\hline
H100 & 2022 & 1000+ & 4th-gen Cores, 80GB HBM3 \\
\hline
\end{tabular}
\end{table}

\textbf{Training Scale:} Training GPT-3 required approximately 3,140 petaflop-days of compute \cite{brown2020language}, \cite{strubell2019energy}, while Google's PaLM utilized two pods of 6,144 TPU v4 chips running continuously for about 50 days \cite{chowdhery2023palm}.

\textbf{Specialized Accelerators:} Companies developed custom ASICs like Google's TPUs, optimized specifically for neural network workloads \cite{jouppi2022tpu}, and startups like Cerebras Systems developed wafer-scale engines to minimize communication bottlenecks \cite{cerebras2021scaling}.

Alongside performance increases, a parallel focus on hardware efficiency has emerged:

\textbf{Architectural Efficiency:} Successive chip generations incorporate improvements to perform AI tasks using less energy, with Google's TPU v4 potentially reducing power usage by up to 30\% for equivalent performance in some workloads \cite{jouppi2022tpu}.

\textbf{Purpose-Built Efficiency:} Algorithmic innovations like EfficientNet achieved high accuracy with significantly fewer parameters and computations, leading to an 8.4x reduction in parameters compared to previous state-of-the-art vision models with similar accuracy \cite{tan2019efficientnet}.

This hardware evolution has enabled AI to transition from research labs into real-world applications:

\textbf{Autonomous Systems:} Tesla's Dojo supercomputer is designed specifically for training neural networks used in self-driving systems \cite{tesla2022ai}.

\textbf{Edge AI:} Neuromorphic chips like IBM's TrueNorth contain a million "neurons" while consuming only about 70 milliwatts, enabling AI tasks in power-constrained environments \cite{merolla2014million}.

\textbf{On-Device Intelligence:} Neural Engines in smartphone processors (e.g., Apple's A16 Bionic with 17 trillion operations per second \cite{apple2022iphone}) allow complex AI tasks to run directly on mobile devices.

\section{LLM Architecture Maturity: The Blueprints of Digital Minds}
The evolution of AI architectures has been pivotal in enabling increasingly sophisticated cognitive abilities, with several key developments transforming how AI models process and understand information.

\subsection{The Transformer Revolution}
The breakthrough came in 2017 with the Transformer architecture \cite{vaswani2017attention}, which dispensed with recurrence in favor of a "self-attention" mechanism. This allowed models to directly model relationships between words regardless of their distance in a sequence, revolutionizing contextual understanding. The Transformer became the foundation for subsequent LLMs, including:

\textbf{BERT} (2018): Used the "encoder" part of the Transformer to learn deep bidirectional representations \cite{devlin2019bert}.

\textbf{GPT Series}: Developed by OpenAI, focused on the "decoder" part of the Transformer for coherent text generation \cite{brown2020language}, \cite{openai2023gpt4}.

\textbf{Architectural Innovations:} As models grew, architectures adapted. Google's Switch Transformer introduced sparsity using Mixture-of-Experts, scaling to over 1.6 trillion parameters while only activating a fraction for any given input \cite{fedus2022switch}. DeepMind's RETRO integrated retrieval mechanisms directly into the architecture \cite{borgeaud2022improving}.

\subsection{Impact on Performance and Adaptability}
These architectural advancements directly translated into quantifiable improvements in AI capabilities:

\textbf{Reasoning Performance:}  Transformer-based architectures enabled a 40.2\% improvement in mathematical reasoning on the GSM8K benchmark when combined with Chain-of-Thought techniques \cite{wei2022chain}.

\textbf{Enhanced Understanding:} Models like BERT demonstrated 10-15\% improvements over previous architectures on natural language understanding benchmarks such as GLUE \cite{devlin2019bert}.

\textbf{Multimodal Integration:} Architecture evolution enabled seamless integration of text with other modalities. Models like PaLM-E \cite{noy2023experimental} achieved 86\% success rates on vision-language robotic planning tasks, compared to 62\% with separate models.

\textbf{Safety Improvements:} Architectural refinements in models like Claude, combined with Constitutional AI techniques, reduced harmful outputs by approximately 20\% compared to earlier models \cite{bai2022constitutional}, \cite{grantham2023autogpt}.

\subsection{Expanding Context Windows}
A crucial parameter of LLM capability is the context window--the maximum amount of information the model can consider at once. Larger windows enable handling longer documents, maintaining conversation history, and complex reasoning across extensive information:

\begin{table}[ht]
\caption{Evolution of LLM Context Windows}
\label{tab:context-windows}
\setlength{\tabcolsep}{1.5pt}
\footnotesize
\begin{tabular}{|l|c|l|l|}
\hline
\textbf{Model} & \textbf{Year} & \textbf{Tokens} & \textbf{Approx. Length} \\
\hline
Early Transformers & 2017-19 & 512-1024 & $\sim$750 words/3 pages \\
\hline
GPT-3 \cite{brown2020language} & 2020 & 2-4K & $\sim$3K words/12 pages \\
\hline
GPT-4 \cite{openai2023gpt4} & 2023 & 8-32K & $\sim$25K words/100 pages \\
\hline
Claude \cite{anthropic2023claude21} & 2023 & 200K & $\sim$150K words/600 pages \\
\hline
Gemini 1.5 \cite{googledeepwind2024gemini15} & 2024 & 1M & $\sim$750K words/3000 pages \\
\hline
\end{tabular}
\end{table}

\begin{figure}[t]
    \centering
    \includegraphics[width=\columnwidth]{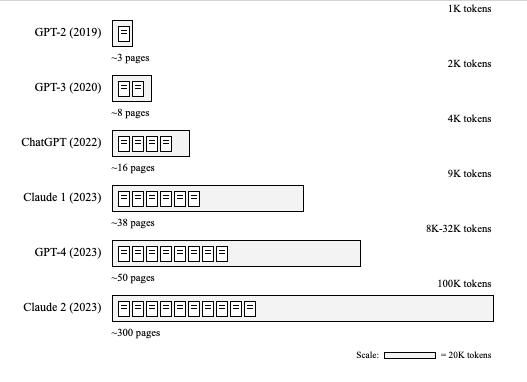}
    \caption{Context Window Size Evolution Visualization}
    \label{fig:cot}
    \end{figure}

This expansion faces challenges including quadratic computational costs in standard Transformers, attention dilution (the "lost in the middle" phenomenon \cite{liu2023lost}), and training complexity. Despite these challenges, larger context windows are fundamental to enabling more complex, knowledgeable, and contextually aware agent behavior.

\section{Expanding AI Capabilities with External Tools: Reaching Beyond the Model}
The next leap towards true AI agency involved equipping models with the ability to interact with the outside world--to use external tools, access real-time data, and take actions beyond simply generating text.

\subsection{Tool Integration and Agent Capabilities}
AI models can now call upon external software applications, APIs, or databases to fulfill user requests or achieve goals, transforming them from passive knowledge repositories into active participants:

\textbf{Information Retrieval:} Models like DeepMind's WebGPT can perform web searches, improving factual accuracy on question-answering tasks by 18\% compared to models without web access \cite{nakano2021webgpt}.

\textbf{Task Execution:} Autonomous agent frameworks like AutoGPT \cite{grantham2023autogpt} can break down complex goals into sub-tasks and autonomously use tools like web search, file system access, and code execution.

\textbf{Plugin Ecosystems:} Platforms like OpenAI's ChatGPT implement plugin systems \cite{openai2023gpt4} enabling functionalities like booking travel, ordering groceries, performing complex calculations, analyzing data, and interacting with e-commerce platforms.

The benefits of tool integration include vastly expanded capabilities, increased accuracy through grounding in real-time data, and automation of complex workflows. GitHub Copilot, for instance, reportedly accelerated coding tasks by up to 55\% in some studies \cite{noy2023experimental}.

However, challenges remain: managing the complexity of interactions between LLMs and multiple tools, addressing security risks from external system access, mitigating additional latency and costs, and training models to effectively use available tools.

\subsection{Benefits and Challenges of Tool Integration: Power and Pitfalls}

\textbf{Benefits:} 
\begin{itemize}
\item \textbf{Expanded Capabilities:} Tool integration allows models to perform tasks beyond their training, such as real-time data retrieval, complex calculations, and multi-step workflows.
\item \textbf{Increased Accuracy:} By grounding responses in real-time data, models can provide more accurate and relevant information.
\item \textbf{Automation of Complex Workflows:} Tools enable models to automate intricate tasks, such as booking travel or managing schedules, enhancing productivity.
\item \textbf{Increased Efficiency:} By automating tasks and integrating with existing systems, models can significantly reduce the time and effort required for complex workflows.
\end{itemize}
\textbf{Challenges:}
\begin{itemize}
    \item \textbf{Error Management:} Tool integration failures can increase task error rates by ~10\% if not properly managed \cite{sevilla2022compute}
    \item \textbf{Security Vulnerabilities:} External system access introduces potential security risks.
    \item \textbf{Performance Impact:} Each tool call adds latency and computational cost
    \item \textbf{Tool Selection:}  Effective use requires sophisticated planning capabilities
    \end{itemize}

Tool integration represents a critical advancement in AI's evolution from passive text generation to active digital agency. It establishes the foundation for versatile AI agents capable of achieving complex goals through a combination of reasoning and interaction with external systems.

\section{AI Agents: The Culmination -- Or the Final Generation?}
The convergence of advanced prompting, sophisticated training, powerful hardware, elegant architectures, and tool integration has created a new class of AI systems--AI agents that can perceive their environment, make decisions, take actions, and learn from outcomes to achieve specific goals.

\begin{figure}[t]
    \centering
    \includegraphics[width=\columnwidth]{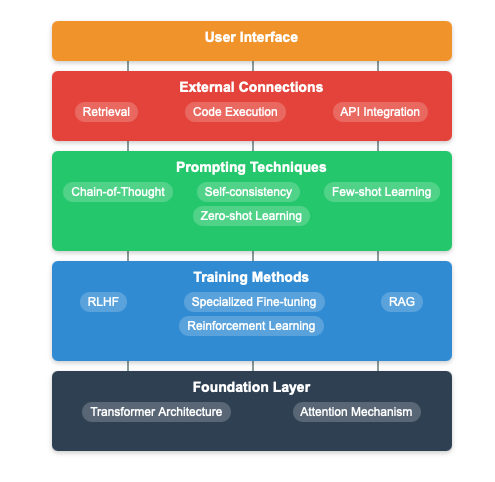}
    \caption{AI Agent Architecture}
    \label{fig:cot}
    \end{figure}

\subsection{The Emergence of Integrated Agents}
Modern AI agents seamlessly integrate multiple capabilities:

\begin{itemize}
\item \textbf{Sophisticated Understanding:} Built on Transformer-based architectures and trained on vast datasets, they possess deep understanding of language, context, and multi-modal information.
\item \textbf{Complex Reasoning:} Techniques like Chain-of-Thought help them break down tasks and plan sequences of actions.
\item \textbf{Access to Knowledge:} RAG techniques and external tools provide real-time, specific information.
\item \textbf{Ability to Act:} Through plugins and APIs, they can execute actions in the digital world.
\item \textbf{Goal-Oriented Behavior:} Unlike passive chatbots, agents pursue multi-step goals and adapt strategies based on intermediate results.
\end{itemize}

Examples include ChatGPT with plugins/code interpreter, Google's Gemini, and specialized research agents like those reported by Anthropic to autonomously design experiments leading to novel chemical discoveries \cite{anthropic2023claude}. Adept AI is building agents designed to learn and execute complex software workflows across enterprise applications \cite{adept2022introducing}.

\subsection{Societal Implications and Challenges}
The rise of capable AI agents carries profound implications across virtually every sector:

\begin{table}[ht]
\caption{Benefits \& Risks of Advanced AI Agents}
\label{tab:ai-implications}
\setlength{\tabcolsep}{1.5pt}
\footnotesize
\begin{tabular}{|l|p{3.7cm}|p{3.7cm}|}
\hline
\textbf{Domain} & \textbf{Potential Benefits} & \textbf{Key Challenges} \\
\hline
Healthcare & Improved diagnostics, drug discovery, personalized care \cite{anthropic2023claude} & Privacy concerns, liability issues, over-reliance \\
\hline
Business & Workflow automation, logistics optimization, customer service & Job displacement (47\% of US jobs at risk) \cite{frey2017future} \\
\hline
Education & Personalized tutoring, adaptive learning, unlimited expertise & Inequality, reduced human interaction \\
\hline
Science & Accelerated research, hypothesis generation, data analysis & Reproducibility, ownership issues \\
\hline
Society & Accessibility, reduced error, economic growth & Bias \cite{buolamwini2018gender}, accountability gaps \cite{sparrow2007killer} \\
\hline
Environment & Resource optimization, climate modeling & Energy costs (GPT-3: 1,287 MWh) \cite{strubell2019energy} \\
\hline
\end{tabular}
\end{table}

\section{The Acceleration: Intelligence Doubling Every Six Months?}
Perhaps the most astonishing aspect of the AI agent narrative is the rate at which capabilities are improving. Independent research from organizations like Epoch AI suggests that AI performance on diverse benchmark suites is effectively doubling approximately every 5.9 months \cite{sevilla2022compute}--a roughly four-fold increase in capability every year.

\begin{table}[ht]
\caption{Accelerating AI Performance (2020-2023)}
\label{tab:ai-acceleration}
\setlength{\tabcolsep}{2pt}
\small
\begin{tabular}{|l|c|c|c|l|}
\hline
\textbf{Benchmark} & \textbf{GPT-3} & \textbf{GPT-4} & \textbf{Gain} & \textbf{Notes} \\
\hline
MMLU & 25-45\% & 86\%+ & 2-3x & Beats humans \\
\hline
GSM8K & 17.7\% & 92\% & 5x & With CoT \\
\hline
HumanEval & 29\% & 67\% & 2.3x & Code gen \\
\hline
LSAT & N/A & 90\% & N/A & Near human \\
\hline
\end{tabular}
\end{table}

This blistering pace is driven by a virtuous cycle of hardware advancements, architectural innovations, improved training techniques, expanding data availability, algorithmic refinements, competitive pressures, and even automated discovery processes like Neural Architecture Search \cite{zoph2017neural}.

If capabilities continue to double every six months, we may see AI systems surpassing human intelligence across most cognitive domains within the next decade. This rapid acceleration underscores the urgency of addressing ethical, societal, and safety challenges while the potential benefits remain immense.

\section{Conclusion: Navigating the Dawn of the Agentic Age}
The journey from the Logic Theorist's rigid proofs \cite{newell1957empirical} to today's fluid, tool-wielding AI agents \cite{openai2023gpt4}, \cite{hassabis2023gemini} represents a fundamental shift in the nature of intelligence on Earth. The confluence of breakthroughs in prompting, training, hardware, architecture, and tool integration has created a new generation of AI with capabilities that increasingly mirror, and in specific domains exceed, human cognition.

These AI agents, potentially the "final generation" of intelligence before a possible singularity or plateau, exhibit remarkable capabilities. The evidence is stark: performance on complex benchmarks like MMLU soaring from ~25\% to over 86\% in under three years, intelligence metrics doubling roughly every six months, and tangible applications transforming industries from scientific discovery to healthcare diagnostics.

Yet this ascent into the agentic age presents profound ethical and societal challenges:

\begin{itemize}
\item \textbf{Amplified Inequity:} Biased algorithms risk deepening societal divides \cite{buolamwini2018gender}.
\item \textbf{Economic Disruption:} Widespread automation could significantly impact employment patterns \cite{frey2017future}.
\item \textbf{Resource Consumption:} The energy footprint of large models clashes with sustainability goals \cite{strubell2019energy}.
\item \textbf{Accountability Gaps:} Autonomous agents blur lines of responsibility \cite{sparrow2007killer}.
\item \textbf{Unforeseen Consequences:} Rapid progress increases the risk of unintended outcomes.
\end{itemize}

The future unfolding is not predetermined by the technology itself, but by our choices in development and deployment. AI agents hold potential for unprecedented human flourishing--curing diseases, solving climate change, personalizing education--but could equally exacerbate inequality, erode trust, and concentrate power if guided by narrow interests or insufficient foresight.

With intelligence potentially quadrupling annually, we are active architects of a new reality. The pace and power of this final generation demand not just technical mastery but deep wisdom, careful deliberation, and globally coordinated efforts to steer towards beneficial outcomes for all humanity. Our legacy will be defined by how we navigate the dawn of the agentic age.

\bibliographystyle{IEEEtran}

\end{document}